# Automatic Discovery of Multi-perspective Process Model using Reinforcement Learning

Sunghyun Sim, *Member, IEEE,* Ling Liu, *Fellow, IEEE* and Hyerim Bae, *Member, IEEE*

**Abstract**—Process mining is a methodology for derivation and analysis of processes models based on the event log. When process mining is employed to analyze business processes, the process discovery step, the conformance checking step, and the enhancements step are repeated. If a user wants to analyze a process from multiple perspectives (such as activity perspectives, originator perspectives, and time perspectives), the above procedure, inconveniently, has to be repeated over and over again. Although past studies involving process mining have applied detailed stepwise methodologies, no attempt has been made to incorporate and optimize multi-perspective process mining procedures. This paper contributes to developing solution approach to this problem. First, we propose an automatic discovery framework of a multi-perspective process model based on deep Q-Learning. Our Dual Experience Replay with Experience Distribution (DERED) approach can automatically perform process model discovery steps, conformance check steps, and enhancements steps. Second, we propose a new method that further optimizes the experience replay (ER) method, one of the key algorithms of deep Q-learning, to improve the learning performance of reinforcement learning agents. Finally, we validate our approach using six real world event datasets collected in port logistics, steel manufacturing, finance, IT, and government administration. We show that our DERED approach can provide users with multi-perspective, high-quality process models that can be employed more conveniently for multi-perspective process mining.

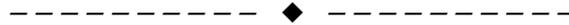

## 1 INTRODUCTION

PROCESS mining [1] is a relatively new research field located between business process modeling and business process analysis and between data mining and operational intelligence [2]. During the past twenty years, process mining has developed rapidly, and many researchers have been continually studying process discovery algorithms that can automatically represent processes from event logs [3]. As a result of this research, various methods for process discovery algorithms have been developed, such as Heuristic Miner (HM)[5], Fuzzy Miner (FM) [6], and Inductive Miner (IM) [7].

Although good process model discovery algorithms have been developed through various studies, the existing approach has the following two limitations.

1) First, the process discovery algorithm does not always derive a process model that expresses the event log well. For this reason, it is necessary to repeat the process discovery, conformance check, and process model enhancement in the process mining area [8]. However, this can sometimes be tedious and unnecessary. In fact, if the purpose of the analysis is clear, the work of discovery, conformance checking, and enhancement of the process model from a single point of view can be completed in one simple cycle. On the other hand, if the purpose of the analysis is not clear, it may be necessary to repeat the cycle of mapping various event data columns to an event log, and then discover the process model, perform conformance checks, and improve it.

2) Second, even if it is desirable to simultaneously derive a process model from various viewpoints based on the purpose of the analysis, the process model is derived only from a certain activity. This is due to the fact that the previously developed process discovery algorithms were designed to derive a process model based on an activity label, which has been selected before running the process discovery algorithm. If we want to derive a process model from various viewpoints, it is inconvenient to map the desired column from the event data to the activity label and then execute the process discovery algorithm again.

While the two above-noted limitations do not prevent the application of process mining, they can be quite inconvenient, and sometimes, there may be problems that require taking additional, otherwise-irrational steps.

The procedure of deriving a multi-perspective process model entails finding a better process model by iterating a set of actions. One great way to solve all of these problems is to

- *Sunghyun Sim is with Industrial Management & Big Data Engineering Major, Division of Industrial Convergence System Engineering, Dongeui University. 176 Eomgwang No, Gaya Dong 24, Busanjin Gu, 47340, Busan, South Korea,, Email: ssh@deu.ac.kr*
- *Ling Liu is with the College of Computing, Georgia Institute of Technology, 266 Ferst Drive, Atlanta, 30332-0765, GA, USA, Email: lingliu@cc.gatech.edu*
- *Hyerim Bae is with a major in Industrial Data Science & Engineering, Department of Industrial Engineering, Pusan National University, 30-Jan-jeon Dong, Geum-Jeong Gu, 609-753, Busan, South Korea, Email: hrbae@pusan.ac.kr (Corresponding author).*

apply reinforcement learning. Reinforcement learning trains an agent to select an action suitable for a specific situation, and it shows good performance with the problem of selecting the optimal action in a fixed environment, such as games [9]-[11] and autonomous driving [12]-[14].

This paper proposes a process discovery technique based on Deep Q-Network (DQN) [15] that can automatically derive the optimal process model by automatically performing process discovery, conformance checking, and enhancement. In addition, this paper proposes a novel experience replay (ER) method called Dual Experience Replay with Experience Distribution (DERED) to improve the learning performance of reinforcement learning agents.

This research offers three benefits to overcome the limitations of the existing research. First, it provides a DQN-based automatic optimal process discovery algorithm that can automatically derive an optimal process model by learning a series of procedures (process model discovery, conformance checking, and enhancement). Second, this research proposes a convenient process discovery algorithm that can simultaneously provide a process model from various points of view, in contrast to its provision from only a single perspective. Finally, a new ER method is introduced (the DERED method) to improve the learning performance of the agent of reinforcement learning that automatically derives the process model.

The remainder of this paper is organized as follows: Section II conducts a literature review. Section III introduces the proposed method for the automated discovery of a multi-perspective process model. Section IV verifies the proposed method using six real-world event logs. Section V summarizes and concludes the discussion.

## 2. BACKGROUND AND RELATED WORK

### 2.1 Process Mining and Process Discovery

Process mining consists of a series of phases, namely process discovery, conformance checking, and model enhancement [16]. The process discovery phase entails the abstraction of the process model from the event log, while conformance checking is a procedure for verifying how well the derived model represents the event log. Then, based on the result of conformance checking, work is done to enhance the process model. The process discovery phase is the basis of all processes.

Research related to process model discovery began with the work of Cook et al [17]. This was the first research on the discovery of process models on FSM (Finite State Machine) from the event log. Afterward, a method of analyzing the process model by a mathematical approach based on several Petri nets such as $\alpha$-miner was proposed [18]. However, there were disqualifying disadvantages, such as vulnerability to actual noise and poor expression of short loop.

Workflow Net (WF-Net) was proposed by Van der Aalst et al. [19] to derive the process model from the execution event log and solve the shortcomings of $\alpha$-miner thereby. In this study, to find a process model in the executed event log, a causal dependency concept based on log-based ordering relations was proposed.

Weijters et al. [20] proposed using frequency to follow the relations in the log according to the log-based ordering relations proposed by van der Aalst et al. [19], and also proposed heuristic miner (HM), one of the most popular process discovery algorithm so far. This algorithm calculates the dependency matrix and considers the frequency between two events $e_i$, and $e_j$ of the following relationships to obtain a casual network. The dependency matrix can be calculated using the equation [20]

$$Dep(e_i, e_j) = \frac{\left| e_i >_L e_j \right| - \left| e_j >_L e_i \right|}{\left| e_i >_L e_j \right| + \left| e_j >_L e_i \right| + 1} \tag{1}$$

where $|e_i >_L e_j|$ is the number of occurrences of $e_j$ directly following $e_i$. The value of $Dep(e_i, e_j)$ is always between -1 and 1; a high value of $Dep(e_i, e_j)$ results from a high number of $|e_i >_L e_j|$ and a low number of $|e_j >_L e_i|$, which indicates that there is a high likelihood of a relation in which $e_i$ is followed by $e_j$. On the other hand, a low value of $Dep(e_i, e_j)$ suggests no dependency between $e_i$ and $e_j$.

Another traditionally popular process model discovery algorithm is fuzzy miner [FM], developed by Günther and Van der Aalst [21]. It aims to derive a simpler and easier-to-see process model by summing and omitting activities and edges according to significance and correlation. Significance means the relative importance of an activity and edge. It is generally judged that the more frequently observed activities and edges are relatively more important. Correlation refers to the degree of association between two activities. If a particular activity occurs and then another particular activity occurs frequently as well, the correlation between the two activities is considered to be high. Using these two measures, the algorithm calculates the relative signification value for filtering edges and the utility value for filtering activity. The relative signification can be calculated using the equation

$$Rel(e_i, e_j) = \frac{1}{2} \times \frac{Sig(e_i, e_j)}{\sum_{e_x \in \sigma} Sig(e_i, e_x)} + \frac{1}{2} \times \frac{Sig(e_i, e_j)}{\sum_{e_x \in \sigma} Sig(e_x, e_j)} \tag{2}$$

where $Sig(e_i, e_j)$ means the frequency for two specific events, and $\sum_{e_x \in \sigma} Sig(e_i, e_x)$ means the sum of the frequency of occurrence between two events $(e_i, e_x)$ belonging to all traces. The utility can be calculated using the equation

$$Util(e_i, e_j) = ur \times Sig(e_i, e_j) + (1 - ur) \times Cor(e_i, e_j) \tag{3}$$

where $Cor(e_i, e_j)$ means the correlation for two specific events, and $ur$ is a parameter between 0 and 1. FM can be a good solution if the event log itself is so complicated that the found process model is too complex, as in the case of the spaghetti model. This algorithm is a suitable approach for event logs with complex structures, such as spaghetti process models [22].

Recently, a new process discovery algorithm called inductive miner (IM) was proposed by Leemans et al [23]. This algorithm represents a new approach to process model discovery that uses a process tree from an event log to extract minimal information. The information provided in this way is known to be able to maintain good quality characteristics while simplifying the complexity of the process model. Also, IM offers polynomial time complexity, which keeps the computational cost viable. IM is currently the most widely used process discovery algorithm, as the process model guarantees soundness, and the quality of the model is normally good [24].

## 2.2 Event Log

One of the most common ways to analyze event logs is process mining [25]. Process mining techniques are used to ex-tract meaningful knowledge from event logs generated from information systems' execution. An event log corresponds to both structured data and semi-structured data. It consists of a case, an activity, a resource, a timestamp, and various attributes.

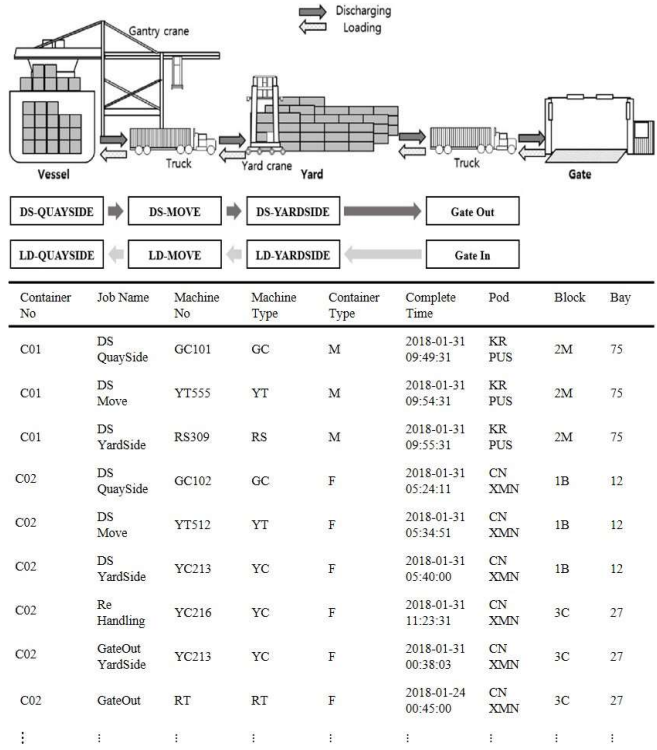

Fig. 1. Structure of environment for discovery of multi-perspective process model

Fig. 1 provides an example of a process execution log compiled from a container-handling process. `C01`, `C02`,`C03`, `C04`, and `C05` are container identifiers, and at the same time, each of them is called case ID. Each row in the Event Log represents an event with a case identifier, an activity label, a time stamp, and originator information.

## 2.3 Experience Replay Algorithm

In this paper, we use deep reinforcement learning for the process discovery using process mining techniques. The Experience Replay (ER) [26] algorithm is the most widely used technique in deep reinforcement learning algorithms based on off-policy, which enables learning even if the learning policy and the action policy are not necessarily the same.

The ER algorithm was first proposed by Lin [27]. With ER, the agent's experience at each time step is stored in a buffer called replay memory. At that time, the agent's experience is defined as a tuple: $e_t = (s_t, a_t, r_{t+1}, s_{t+1})$. This tuple contains the state of the environment, $s_t$, the action, $a_t$, taken from state $s_t$, the reward, $r_{t+1}$, given to the agent at time $t+1$ as a result of the previous state-action pair $(s_t, a_t)$, and the next state of the environment, $s_{t+1}$. This tuple indeed gives us a summary of the agent's experience at time $t$. This replay memory dataset will be randomly sampled from the memory buffer for network training.

The ER algorithm is again gaining attention after Mnih et al.'s study [15]. In earlier studies, reinforcement learning had used agents' past experiences ($e_t = (s_t, a_t, r_{t+1}, s_{t+1})$) as training data for Q-function estimation. However, there was a strong correlation between such experiences. For this reason, it was known that if a nonlinear approximator, such as a neural network, was used to estimate the Q-function, it was difficult to obtain a good Q-function. What Mnih et al. did was to reintroduce the experiential replay technique to solve this problem, reduce the correlation between the learned data, and successfully estimate the Q-function.

Schaul et al. [28] introduced the Prioritized Experience Replay (PER) algorithm with an improved ER algorithm to improve training performance. Unlike the ER algorithm performing uniform sampling in the replay buffer, PER prioritizes the experiences, weighting the samples to extract relatively meaningful experiences more often for training [29]. In PER, the experience of increasing the difference between the TD target and the actual Q-value is first sampled to be learned. At this time, extracting only high-priority experiences can lead to overfitting. To prevent overfitting, PER tries to take samples even in low-priority environments. In particular, 57 different experiments showed that PER performance was almost three times higher than that of the ER method, which performs sampling from a uniform distribution.

Later, several researchers started to study the importance of replay buffer size. In much of the previous research, the replay buffer had been set to a default capacity. It is now known that this difference does not matter if the ER is robust [30]. However, the replay buffer size can significantly reduce the learning speed and the quality of the learning results. First of all, if the replay buffer is too small, it often happens that the replay buffer is rarely or never used. Secondly, if the replay buffer is too large, the batched samples are not correlated, but a problem arises when the agent learns in a modern environment after a long time. In order to solve this problem, Zhang and Sutton [31] proposed a new ER algorithm called Combined Experience Replay (CER) that combines the existing ER method with the online learning method and that is robust to memory buffer size. Compared with the conventional method, this method is relatively insensitive to the replay buffer size, and demonstrated experimentally that learning is performed well with a smaller amount of data.

## 3 PROPOSED METHODS FOR AUTOMATIC DISCOVERY OF MULTI-PERSPECTIVE PROCESS

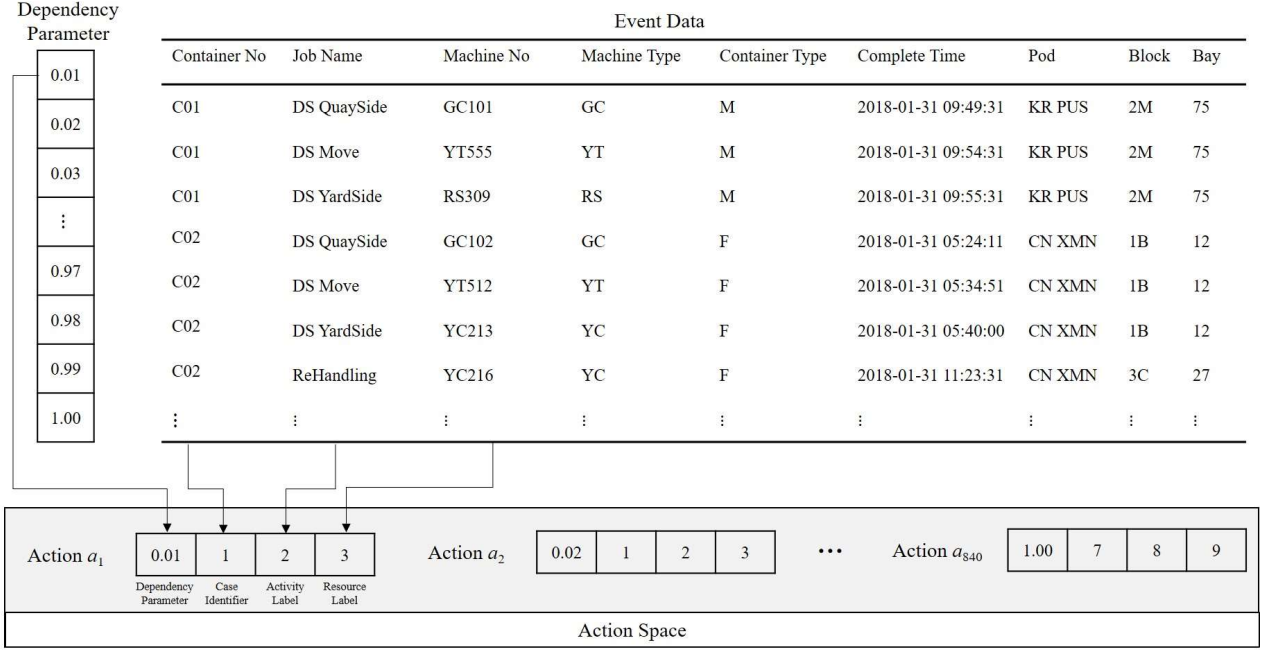

| Dependency Parameter | Event Data | | | | | | | | |
|---|---|---|---|---|---|---|---|---|---|
| | Container No | Job Name | Machine No | Machine Type | Container Type | Complete Time | Pod | Block | Bay |
| 0.01 | C01 | DS QuaySide | GC101 | GC | M | 2018-01-31 09:49:31 | KR PUS | 2M | 75 |
| 0.02 | C01 | DS Move | YT555 | YT | M | 2018-01-31 09:54:31 | KR PUS | 2M | 75 |
| 0.03 | C01 | DS YardSide | RS309 | RS | M | 2018-01-31 09:55:31 | KR PUS | 2M | 75 |
| ⋮ | C02 | DS QuaySide | GC102 | GC | F | 2018-01-31 05:24:11 | CN XMN | 1B | 12 |
| 0.97 | C02 | DS Move | YT512 | YT | F | 2018-01-31 05:34:51 | CN XMN | 1B | 12 |
| 0.98 | C02 | DS YardSide | YC213 | YC | F | 2018-01-31 05:40:00 | CN XMN | 1B | 12 |
| 0.99 | C02 | ReHandling | YC216 | YC | F | 2018-01-31 11:23:31 | CN XMN | 3C | 27 |
| 1.00 | ⋮ | ⋮ | ⋮ | ⋮ | ⋮ | ⋮ | ⋮ | ⋮ | ⋮ |

Action $a_1$ | 0.01 | 1 | 2 | 3    Action $a_2$ | 0.02 | 1 | 2 | 3   ⋯   Action $a_{840}$ | 1.00 | 7 | 8 | 9

Dependency Parameter / Case Identifier / Activity Label / Resource Label

Action Space

Fig. 2. Example of defining of action space using dependency parameters and port logistics operation data

## MODEL

This section introduces a methodology for automatically discovering a multi-perspective process model based on reinforcement learning and the DERED method.

### 3.1 Definition of Deep Q-Learning Components

Here, we define the environment for automated process discovery, the action and the state. Further, we define the method of calculating the reward.

The action consists of a tuple of 4 elements. The first element contains information about the parameters used in the process model algorithm. For example, if the agent of the DQN to be trained uses the process model algorithm as the HM, the dependency parameter is used. In this case, the first element of the task is assigned a real number between 0 and 1 in units of 0.01. The remaining elements among the four are related to which column to be selected as the case identifier, activity label, and resource label, respectively, in the event data. The remaining 3 elements are the combination of randomly selecting 3 from the total number of data columns. If there are $n$ columns in the data and $p$ parameters for the discovery of the process model, the total workspace is equal to the number of $p \times {}_nC_3$.

Fig. 2 provides an intuitive description of how to define an action space. From the first element of the action, one of the dependency parameters of HM is defined. At this time, one of the values from the smallest to the largest value of 1 is assigned as the first element of the action. Column numbers corresponding to the case identifier, activity label, and resource label are allocated from the 2nd to 4th elements of the action, respectively. As a result, the first action is defined as a tuple of <0.01,1,2,3>. By repeating this process, the entire action space is defined, and the size of the total action space becomes 840 (100×${}_9C_3$).

For training so that the agent can derive a process model that guarantees various and high fitness, it is necessary to define the states well. For example, when using an HM to derive a process model, the process model that we see as a result is visualized as a dependency graph using the result derived through the dependency matrix. In other words, since the process model is shown through the heuristic discovery algorithm and is expressed based on the dependency matrix, it is desirable to provide information about the state as a dependency matrix to the agent of deep Q-learning. For this reason, we define the state as an HM's dependency matrix. Another advantage we get when we define the dependency matrix as a state is that we can provide the state directly to the network as an input during agent training without any preprocessing. In the DQN, the state is used as an input to learn the policy network and is mainly expressed in the form of an $n$- dimension array. Since the dependency matrix is a 2-dimensional array calculated from the event log through a heuristic discovery algorithm, it has the advantage of being directly used as an input to a policy network. The value of the dependency matrix is calculated using Equation (1).

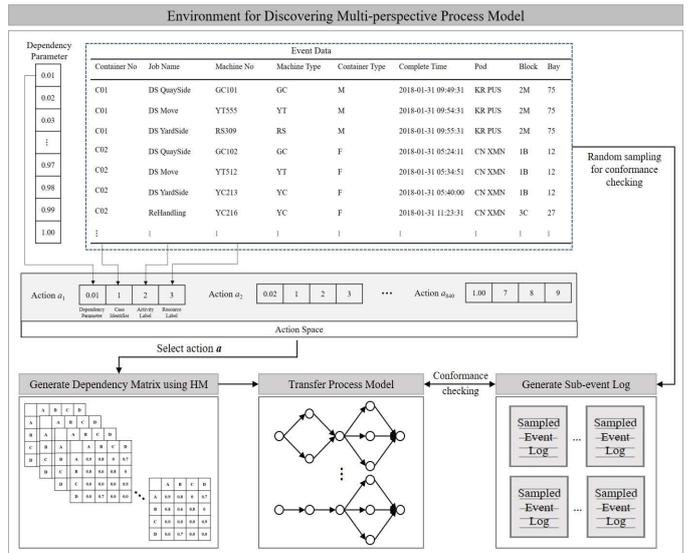

Fig. 3. Structure of environment for discovery of multi-perspective process model

Environment refers to a space in which actions determined by a policy network are performed [32]. In our proposed framework, the environment is defined as a space where an agent automatically discovers a process through an action determined and performs conformance checking. In the environment we developed, the process discovery algorithm uses an HM, and the conformance check measure uses the model fitness.

Fig. 3. shows the structure of the overall environment of deep Q-learning that we use for learning. Finally, the reward function is the fitness value of the process model derived when an action is given. The quality of the process model is intuitively expressed through fitness. The agent has a high reward when taking action to discover a process model with a high fitness value.

### 3.2 Dual Experience Replay with Experience Distortion

In this section, we will introduce a new replay experience method called Dual Experience Replay with Experience Distortion (DERED) for improving learning performance. The experience replay (ER) method [26] is a key technology that helps the Q-function quickly converge using previous experiences, thus helping to determine the exact value of the action taken by the agent [27]. However, if the ER used for Q-function learning is full of experiences that do not satisfy the conditions we want, there is a high possibility that the agent will not be able to accurately evaluate the goodness of the experiences, even if it does an action that satisfies the desired conditions.

The previously proposed ER technique learns a value evaluation Q-function by randomly sampling past experiences performed by agents from replay buffers. In contrast, our study suggests a weighted sampling method between success and failed experiences, suggesting methods that agents can use for learning.

Suppose that we have an experience wherein success and failure are clearly separated. Sometimes the probability of success is very low, and if the sampled experience also does not contain many success cases, it is more likely to learn a biased Q-function, leading to serious problems involving misestimation of the value of the action. To solve this problem, we propose a new ER method, called DERED, of the structure shown in Fig. 4.

The structure of the experience buffer storing the experience uses a double experience buffer structure that can separate and store success and failed experiences, unlike the conventional ER method, that used a single-experience buffer. Here, the successful experience and the failed experience distinguish between success and failure based on whether the agent exceeds or does not exceed the minimum fitness value set by the user when the process model is extracted. We use the over-sampling and under-sampling methods at the same time in the buffer containing the successful experience and the buffer containing the failed experience, respectively, unlike the conventional ER method, which randomly extracts the experience to be used for learning. We want the Q-function to learn failure cases and success cases equally so that they are not biased to one side. To solve this problem, we adjust the two experiences to balance by applying over-sampling to success cases and under-sampling to failed experiences early in learning.

In the sampling step, we succeeded in balancing the two experiences by applying different sampling methods to the two buffers, but there may be an overfitting problem in which certain experiences are over-learned. We solve this problem by providing the concept of experience distortion. Experience distortion is carried out in the following 3-step procedure.

1) **Step1:** A parameter $\lambda$ between 0 and 1 is set, and then generates a distortion ratio $r$ that will cause experience distortion from the uniform distribution $U(0, \lambda)$.

2) **Step2:** The experience to be distorted is extracted from the entire sampled experience by the ratio generated through Step 1.

3) **Step3:** The experience is distorted by randomly replacing the action of the extracted experience with another action

Fig. 5 shows the procedure for performing experience distortion using the action space of Fig. 2 introduced earlier. Suppose that you have 10 experiences sampled here for the training of agents. It is also assumed that after extracting the $\lambda$ between 0 and 1, a distortion ratio $r$ of 0.2 is generated from a uniform distribution $U(0,\lambda)$. As a result, suppose that the 6th sampled experience and the 10th sampled experience are selected for experience distortion. At this time, the two experiences $a_{65}$, $a_{99}$ are randomly replaced with any experience that exists in a total of 840 action spaces. Looking at the bottom of Fig. 5, the action in the sixth experience is composed of $a_{65}$ = <0.85,1,8,3>, but through the experience distortion, it can be seen that it is randomly replaced with $a_{65}$ = <0.90,5,7,3>.

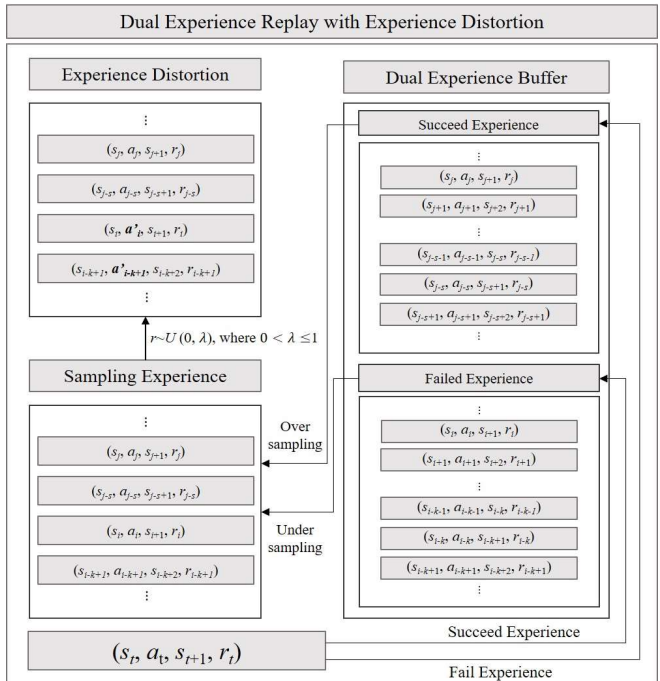

Fig. 4. Structure of environment for discovery of multi-perspective process model

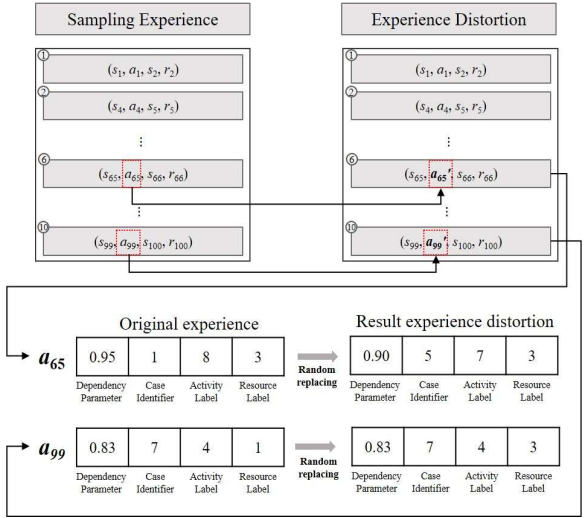

Fig. 5. Structure of environment for discovery of multi-perspective process model

### 3.3 Procedure for Q-function Approximation

This section introduces the learning structure and network structure to approximate the Q-function that evaluates the value of an action. In Mnih's work [15], a CNN-based DQN algorithm was proposed to approximate the Q-function. This algorithm represents the current state as an image and proposes a method to estimate the present value function using CNN. In the proposed method, we also approximate the Q-function using the DQN approach proposed by Mnih et al.

Fig. 6 shows the general structure of network training for approximation of the Q-function. We use this process to converge our Q-function. The detailed learning procedure is as follows.

1) **Step1:** Separating sampled experiences into batches for Q-learning.
2) **Step2:** Putting $s_t$ (current state) and $s_{t+1}$ (next state) of the experience included in each batch into the policy network and target network, respectively.
3) **Step3-1:** Estimating $Q(s_t, a_t; \theta)$ from the policy network. Here, refer to $\theta$, the parameter of the policy network.
4) **Step3-2:** Estimating $r_t + \gamma(\max_{a_{t+1} \in A} Q(s_{t+1}, a_{t+1}; \theta'))$ from the target network. Here, $r_t$ and $\gamma$ refer to the reward and discount factors, respectively. $a_{t+1}$ refers to all possible actions included in the action space, and $\theta'$ refers to the parameter of the target network.
5) **Step4:** Calculating the loss using $Q(s_t, a_t; \theta)$ and $r_t + \gamma(\max_{a_{t+1} \in A} Q'(s_{t+1}, a_{t+1}; \theta'))$ estimated in Step 3-1 and 3-2, respectively. The loss [33]-[35] is defined by the following Equation (4):

$$l = E[r_t + \gamma\left(\max_{a_{t+1} \in A} Q'(s_{t+1}, a_{t+1}; \theta')\right)]^2 \quad (4).$$

6) **Step5:** Optimizing of the parameters $\theta$ of the policy network for each experience using Adam optimizer [39].

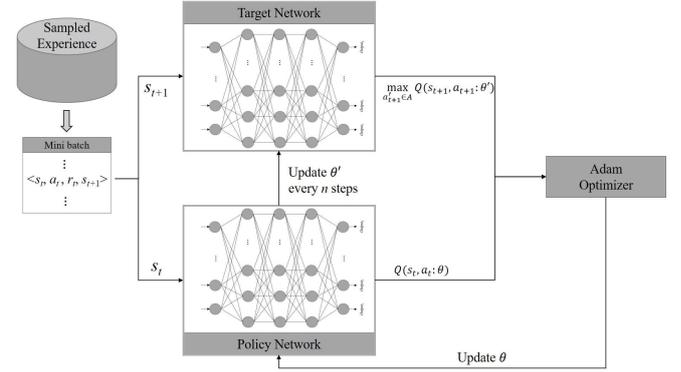

Fig. 6. Structure of environment for discovery of multi-perspective process model

Then, every $n$ steps, parameters $\theta'$ of the target network are updated using parameters $\theta$ of the policy network.

7) **Step6:** Repeating steps 1 through 5 to train the target network to approximate the optimal Q-function.

As for the policy network and target network introduced in Fig. 6, after Mnih's study, the network using the structure of CNN [36] has been the most frequently used [37], [38]. We analyzed previous studies [39]-[41] using CNN as an approximation function and defined CNN as a network. Table I shows the structure of CNN for Q-value approximation and provides a summary of input and output shapes.

TABLE I
SUMMARY OF NETWORK STRUCTURE FOR APPROXIMATION OF Q-VALUE

| Layer Block | Configuration | Input Shape | Output Shape |
|---|---|---|---|
| Convolution Block₁ | Convolution Layer<br>Batch Normalization<br>Activation Function | 128×128×3 | 32×32×32 |
| Convolution Block₂ | Convolution Layer<br>Batch Normalization<br>Activation Function | 32×32×32 | 14×14×64 |
| Convolution Block₃ | Convolution Layer<br>Batch Normalization<br>Activation Function | 14×14×64 | 12×12×64 |
| Fully-connected Block₁ | Fully Connected Layer<br>Batch Normalization<br>Activation Function | 12×12×64 | 1×512 |
| Fully-connected Block₂ | Fully Connected Layer | 1×512 | 1× (Number of Action) |

### 3.4 Deep Q-Learning Framework for Automated Discovery of Multi-perspective Process Model

We now introduce the composition and detailed procedure of our proposed algorithm. The training procedure for the AMPD is performed as follows in seven steps.

1) **Step1: Initializing the state.** The state is represented as the dependency matrix used in Equation (1). The initial state consists of a total dependency value of zero.
2) **Step2: Selecting the action.** The agent selects the action (select case identifier, activity label, resource label) to find a process model that can satisfy a given minimum fitness. Firstly, we choose random epsilon values from a uniform distribution to select the action. Then, we use either exploration by the selected epsilon value (random action selection) or exploitation (action selection using learned policies such as policy network) to select the highly rewarding actions in a current state.

3) **Step3: Observing the new states using the selected action.** Observe the new state by using the action selected through Step2. Then, calculate the dependency matrix using the selected action and store it as a new state.

4) **Step4: Calculating the reward of the new state.** We calculate the fitness of the process model and the deviation of the derived dependency matrix. It uses the sum of the two values as the reward function.

5) **Step5: Collecting experiences and storing them in the replay buffer.** Experiences including current status, action, the new state, and rewards derived through steps 1 to 4 are created and stored in the replay buffer. The experience here is the tuple of four elements introduced earlier. This process is repeated to the extent that the batch size is defined during reinforcement learning.

6) **Step6: Updating network through experience replay.** The data used for policy network and target network training is sampled from the experience's replay buffer. Then, it uses the data consisting of the sampled experiences to perform a network learning process for policy-making.

7) **Step7: Repeating steps 1 to 6 for each episode.** The process from Step 1 to Step 6 is defined as one episode, and once the set episode is completed, one learning process is terminated. If the specified learning process repeats $n$ times, and the learning process consists of $m$ episodes, the entire learning repeats $n \times m$ episodes.

Algorithm I is the pseudo-code of the AMPD algorithm based on DQN and DERED

## 4. EXPERIMENTS

For the experiments validating our approach, we use real-world event logs from 6 different areas. First, we describe

---

**Algorithm I** AMPD based on DQN with DERED

| | |
|---|---|
| **Input:** | The trade-off parameter $\pi$, The distortion rate $\lambda$, The learning rate $\alpha$, Adam hyperparameters $\beta_1, \beta_2$ |
| **Output** | Optimal Q-function |

1  Initialize successful experience buffer $D_s$ and initialize failed experience buffer $D_f$
2  Initialize policy network parameter $\theta_Q$ and initialize target network parameter $\theta_{\hat{Q}}$
3  $\theta_{\hat{Q}} \leftarrow \theta_Q$ // Updating target network parameter to policy network parameter
4  **for** $e \leftarrow 1$ to $E$ **do** //E is the number of epochs
5     initialize state $s$ // s is dependency matrix and initial state is matrix with zero value
6     loss $\leftarrow$ []
7     **for** $t \leftarrow 1$ to $T$ **do** // T is the number of trials
8        randomly sampling probability $p \sim U(0,1)$
9        **if** $p < \varepsilon$ **then** // $\varepsilon$ is $\varepsilon$ -greed policy
10          $a_t \leftarrow$ randomly select action
11       **else**
12          $a_t \leftarrow argmax_a Q(s_t, a_t; \theta_Q)$
13       $s_{t+1} \leftarrow a_t(s_t)$
14       $r_t \leftarrow mean(s_{t+1}) + std(s_{t+1})$
15       $f \leftarrow fitness(s_{t+1})$
16       **If** $f > \lambda$ **then**
17          $D_s \leftarrow < s_t, a_t, s_{t+1}, r_t >$ //storing successful experience
18       **else**
19          $D_f \leftarrow < s_t, a_t, s_{t+1}, r_t >$ //storing failed experience
20       $D \leftarrow combine(oversampling(D_s), undersampling(D_f))$ //Experience sampling
21       **for** $k \leftarrow 1$ to $|D|$ **do**
22          Randomly sampling $r \sim U(0,1)$
23          **if** $r < \lambda$ **then**
24             $a_t' \leftarrow$ randomly selected action
25             $D[k] \leftarrow < s_t, a_t', s_{t+1}, r_t >$ // Exploration
26          **else**
27             $D[k] \leftarrow < s_t, a_t, s_{t+1}, r_t >$ // Exploitation
28       **for** $j \leftarrow 1$ to $|D|$ **do** // Updating Q-function
29          $y_t = \{r_t + \gamma \max_{a'} \hat{Q}(s_{t+1}, a'; \theta_{\hat{Q}})$ , $f > \pi$  , otherwise
27          loss $\leftarrow Adam(\nabla_\theta \sum_{i=1}^{|D|} loss$ with $\theta_Q, \alpha, \beta_1, \beta_2)$
28       **end**
29       $\theta_{\hat{Q}} \leftarrow \theta_Q$
30    **end**
31 **end**
32 Optimal Q-function $\theta_Q$

the datasets used for the experiments. Then, we introduce the experimental design. Finally, we show the comparison results of the experiments.

### 4.1 Dataset Description

Our experiments use six event datasets (from TL1 to TL6) collected in port logistics, steel manufacturing, finance, IT, and government administration. Notice that TL3 to TL6 are publicly available event data used in the Business Process Intelligence Challenge of 2012 to 2016 (BPI 2012 to BPI 2016). The detailed description of each event dataset is as follows.

\

TABLE II
SUMMARY OF EXPERIMENTAL DATASETS DESCRIPTION

| Data | Number of Cases | Number of Events | Data descript |
|------|------|------|------|
| TL1 | 13,954 | 55,004 | Logistic processes within a Korean port company |
| TL2 | 77,405 | 1,048,575 | Steel manufacturing processes within a Korean company |
| TL3 | 13,087 | 13,087 | Bank lending process in a Netherlands company |
| TL4 | 7,554 | 65,533 | Incident management process within Volvo |
| TL5 | 46,616 | 644,738 | ITIL process within Rabobank Group |
| TL6 | 1,199 | 52,217 | Building permit processes from five Dutch municipalities |

### 4.2 Experimental Method

We want to check the following two things through experiments in this section. First, we want to check whether the newly proposed ER method can guarantee superior learning performance compared with the existing ER method. Second, we want to check if our proposed deep Q-learning for multi-perspective process discovery works well. If it works fine, we should be able to generate several process models that can satisfy the specific fitness. To achieve the above two objectives, we conduct two experiments using the six experimental datasets.

Learning of deep Q-learning for experimentation proceeds as follows. One episode ends when the agent creates 50 process models. Each episode is repeated 100 times to proceed with learning. The minimum fit to determine success and failure is defined as 0.7. If the fitness value of the process model created by the agent is more than 0.7, it is considered a success case. On the other hand, if the fitness value is less than 0.7, it is considered a failure case. This process is repeated 30 times. According to the training procedure described above, we perform three experiments comparing the performances between the two experience replay methods (ER, PER) and the proposed method (DERED).

1) **EXP-I:** An experiment to identify ER methods helping agents to learn in a direction where they can receive higher rewards. To compare this, we compare the total average score. The score is the total reward the agent receives in one episode. First, we measure the average score for the total of 30 tests the agent received. We want to use this metric to see if our learning is progressing towards more rewards.

2) **EXP-II:** Experiments to identify ER methods that contribute to learning in the direction in which agents can discover high-quality process models. They measure the average fit of the 50 process models that the agent discovered in each episode. The average of the fitness values collected over the 30 iterations is calculated. We try to use this measure to see if there are any changes in the quality of the process model that the agent discovered during training.

3) **EXP-III:** An experiment to identify ER methods that contribute to learning in a direction that better meets the minimum fitness set by the user. It measures the number of process models with a fitness value of 0.7 or more out of the 50 process models the agent discovered in each episode. It is collected repeatedly over 30 times. It is determined whether it is possible to automatically generate several types of process models that satisfy the minimum fit set by the user as learning progresses.

EXP-I to EXP-III are conducted using a stand-alone computer with operating system Windows 10, AMD Ryzen 7 2700 Eight-core Processor (3200Mhz) with 64Gb memory. The model training is done using NVIDIA GeForce RTX 2070-8GB Dedicated GPU Memory with Cuda version 10.2. The model is implemented using PyTorch 1.2.0.

### 4.3 Experiment Results

Fig. 7 shows the total average score of the agent for deriving the automatic process model according to the respective ER method. In the figure, the x-axis refers to the number of the epochs that has been trained, and one epoch is composed of one episode. In one episode, the agent derives a total of 50 process models. The y-axis refers to the total average score acquired by the agent while performing the episode. As can be seen from the two figures below, we can see that the DERED method we proposed has a faster convergence speed than the other two methods. In addition, it shows that as the learning progresses, the agent that automatically derives the process is learning to receive higher rewards.

Table III shows the average of the total score last acquired (100th epochs) by the agent after training is completed. Looking at this result, the reward has been 1.7 to 3 times higher than that of the ER method. It also shows that it has acquired about 1.5 times higher compensation than the known-to-be-excellent PER method. The results in the table prove that our proposed method can learn a better agent. Fig. 8 shows the averages of process model fitness calculated by repeating, 30 times, the average fitness for 50 process models discovered by the agent for each episode. As in Fig. 7, the x-axis refers to the number of epochs.

Fig. 8 shows the averages of process model fitness calculated by repeating, 30 times, the average fitness for 50 process models discovered by the agent for each episode. As in Fig. 7, the x-axis refers to the number of epochs. It shows that the quality of the average process model improves as the agent learns, regardless of the ER method. In Fig. 8, we can observe that the average fit of the process model is increasing under all conditions as the training progresses. This means that as learning progresses, the actions taken by the agent are learned in the direction of improving the quality of the process model. In other words, this is an example that shows that our proposed AMPD-algorithm-based deep Q-learning works well.

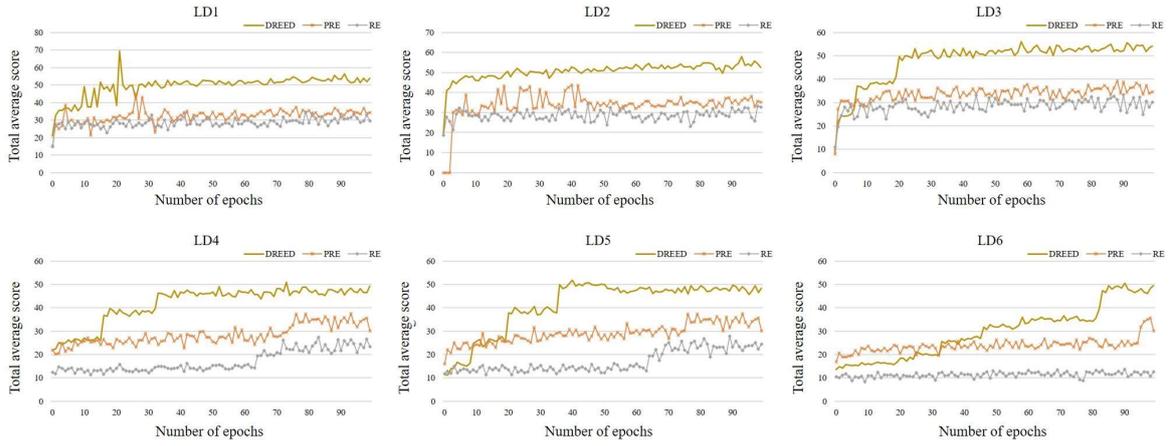

Fig. 7 Result plot of EXP-I with 30 iteration

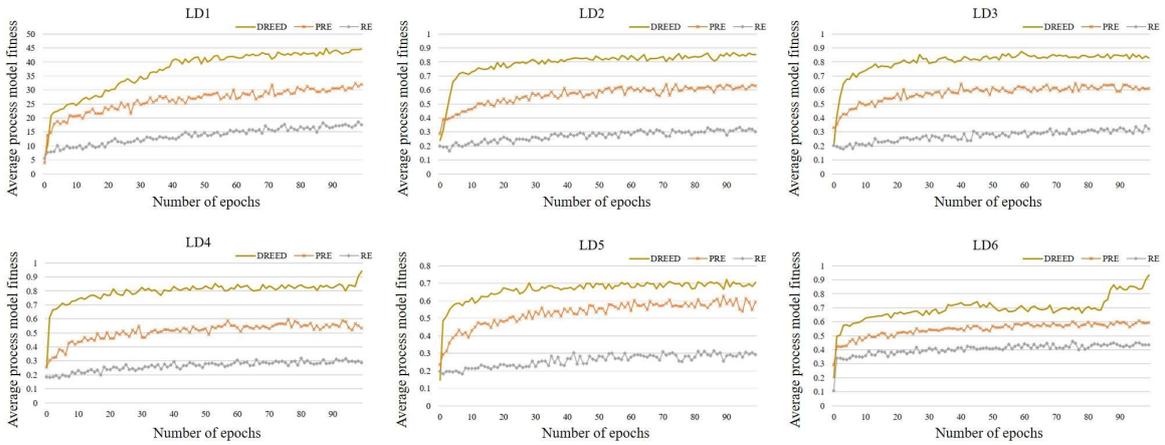

Fig. 8 Result plot of EXP-II with 30 iteration

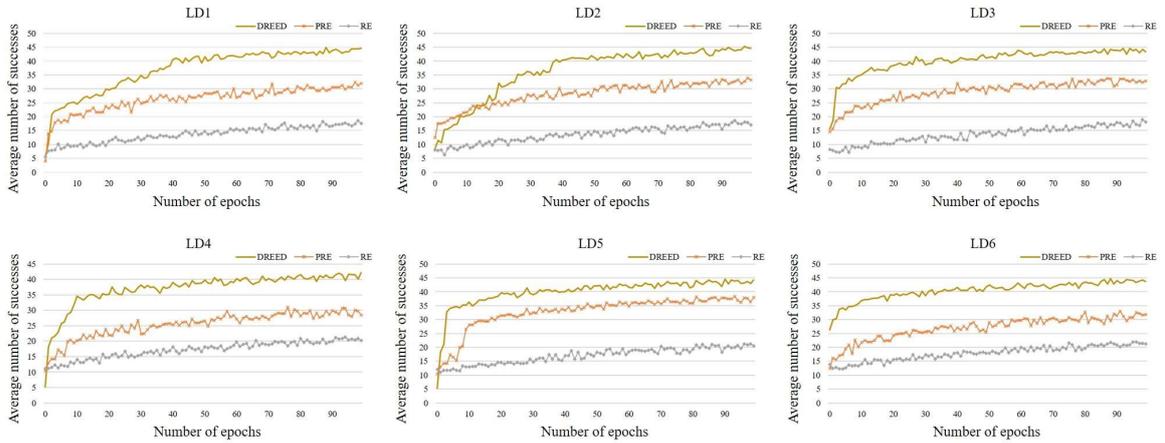

Fig. 9 Result plot of EXP-III with 30 iteration

Comparing the differences among the ER methods, we can see that the DERED method proposed by us learns an agent to derive a better-quality process model. In addition, in all six datasets used in the experiment, the proposed method shows that the quality of the process model discovered by the agent is improved at a rapid rate relative to the other methods.

Table IV shows the average process model fitness values for the last 50 process models ($100^{th}$ epoch) discovered by the agent after training is complete. Looking at these results, the agent trained using the proposed method derived a process model of about 2.1 to 3.1 times higher quality than could the agent derive using the ER method. In addition, it can be seen that a process model of about 1.2 to 1.7 times higher quality than the PER method was derived.

Fig. 9 shows a comparison of how well the agent derives a process model that satisfies the minimum fit set by the user while performing training. It counts how many of the 50 process models derived when performing an episode have a fit higher than 0.7. In most of the experimental results, in the initial epoch, only 5-10 process models out of 50 satisfied the minimum fitness. However, as the learning progresses, the average fitness of the process model that the agent discovers improves; and, it is observed that more process models meeting the minimum fitness criteria are found. In this experiment, the DERED method we proposed still shows the highest learning performance, and it also shows that the quality of the process model found at the time the training is completed is superior to the other two methods.

## 5 DISCUSSION AND CONCLUSION

This paper proposes a new method for automatic discovery of processes by which processes can be derived from multiple perspectives at once using DQN, one of the reinforcement learning techniques. To further optimize DQN, we also develop a effective method for performing three steps (discovery, fitness test, and improvement) of process model). The method is designed to give higher rewards as the agent finds more process models that meet the minimum fit during learning. The most distinctive advantage of this proposed method over the existing process search algorithms is that it simultaneously searches for the process model from multiple viewpoints in the event log.

Another major contribution of this paper is the development of a new experience replay (ER) method to improve the agent's learning performance. A dual-experience buffer is proposed, which separates and stores successful and unsuccessful experiences, to improve the mining efficiency compared to the existing methods that use only a single-experience buffer. To solve the problem of over-learning only failure cases when training agents, success cases and failure cases are sampled in a balanced way and used for training. Finally, the new ER method, called DERED, applies the concept of experience distortion. To evaluate the performance of DERED, a comparison experiment was conducted in which two popular ER methods were applied to six real-world event logs. The experimental results show that the learning performance of the agent using the proposed DERED method was superior with all of the data. Several experiments demonstrated that the agent discovers a higher-quality process model as it performs training.

This research is expected to provide users with multi-perspective, high-quality process models more conveniently in the area of process mining. Furthermore, the newly proposed DERED method is expected to be effectively applied to DQN, which is widely used in reinforcement learning, to contribute to improved learning performance.

## ACKNOWLEDGMENT

This work was supported by a National Research Foundation of Korea (NRF) grant funded by the Korean government (MSIT) (No. 2020R1A2C110229411).

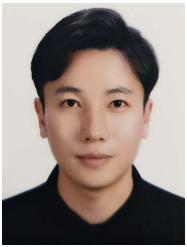

**Sunghyun Sim** received his BS. degree in Statistics from Puan National University, Rep. of Korea, in 2016, and his MS and Ph.D. degrees in Industrial Engineering from Pusan National University, Rep. of Korea in 2021. Since 2022, he has been an assistant professor with Industrial Management & Big Data Engineering Major, Division of Industrial Convergence System Engineering Dongeui University, Rep. of Korea His research interests include data mining; process mining; deep learning; big data analytics for operational intelligence; and process optimization based on the deep-learning method

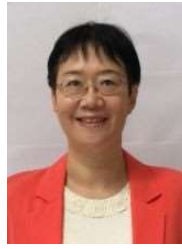

**Ling Liu** is a Full Professor at the College of Computing, Georgia Institute of Technology, Atlanta, GA, USA. She directs the research programs in Distributed Data Intensive Systems Lab (DiSL), examining various aspects of large-scale data-intensive systems. Her current research interests include performance, availability, privacy, security and trust of big data systems, distributed computing, as well as Internet-scale systems and applications. She has published over 300 international journal and conference articles and is a recipient of the Best Paper Award from numerous top venues, including IEEE ICDCS, The World Wide Web Conference, IEEE Cloud, IEEE ICWS, and ACM/IEEE CCGrid. Prof. Liu has served on the editorial boards of over a dozen international journals. Currently, Prof. Liu is the Editor-in-chief of ACM Transactions on Internet Computing (TOIT). Her current research is sponsored primarily by NSF and IBM.

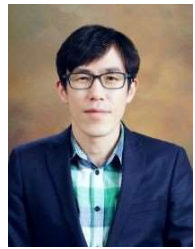

**Hyerim Bae** received his BS, MS, and PhD degrees in Industrial Engineering from Seoul National University, Rep. of Korea. From 2002 to 2003, he worked for Samsung Credit Card Corp., Seoul, Rep. of Korea. Since 2005, he has been a professor with the Department of Industrial Engineering, Pusan National University, Rep. of Korea. He is interested in information system design; cloud computing; business process management systems, and process mining and big data analytics for operational intelligence.
His current research activities include analyzing huge volumes of event logs from port logistics and shipbuilding industries using process mining techniques